\documentclass[10pt,twocolumn,letterpaper]{article}

\usepackage[pagenumbers]{cvpr} 

\usepackage{graphicx}
\usepackage{amsmath}
\usepackage{amssymb}
\usepackage{booktabs}

\usepackage{multirow}
\usepackage{cite}
\usepackage{color}
\usepackage[table]{xcolor}

\usepackage{algorithm}
\usepackage{algorithmic}
\usepackage{amsmath, amsfonts}

\usepackage[pagebackref,breaklinks,colorlinks]{hyperref}

\usepackage[capitalize]{cleveref}
\crefname{section}{Sec.}{Secs.}
\Crefname{section}{Section}{Sections}
\Crefname{table}{Table}{Tables}
\crefname{table}{Tab.}{Tabs.}

\def\cvprPaperID{4776} 
\def\confName{CVPR}
\def\confYear{2022}

\begin{document}
	
	\title{Mix-Teaching: A Simple, Unified and Effective Semi-Supervised Learning Framework for Monocular 3D Object Detection}

	\author{%
		Lei Yang\textsuperscript{1},  Xinyu Zhang\textsuperscript{1}\thanks{Corresponding author: xyzhang@tsinghua.edu.cn},  Li Wang\textsuperscript{1},  Minghan Zhu\textsuperscript{2}, Chuang Zhang\textsuperscript{1},  Jun Li\textsuperscript{1} \\
		\textsuperscript{1}State Key Laboratory of Automotive Safety and Energy,  Tsinghua University \\
		\textsuperscript{2}University of Michigan \\
		\begin{normalsize}${\tt \{yanglei20, zhch20\}@mails.tsinghua.edu.cn; \{xyzhang, wangli\_thu,lijun1958\}@mail.tsinghua.edu.cn}$\end{normalsize} \\
		\begin{normalsize}${\tt minghanz@umich.edu}$\end{normalsize}
	}
	\maketitle
	\begin{abstract}
		Monocular 3D object detection is an essential perception task for autonomous driving. However, the high reliance on large-scale labeled data make it costly and time-consuming during model optimization. To reduce such over-reliance on human annotations, we propose Mix-Teaching, an effective semi-supervised learning framework applicable to employ both labeled and unlabeled images in training stage. Mix-Teaching first generates pseudo-labels for unlabeled images by self-training. The student model is then trained on the mixed images possessing much more intensive and precise labeling by merging instance-level image patches into empty backgrounds or labeled images. This is the first to break the image-level limitation and put high-quality pseudo labels from multi frames into one image for semi-supervised training. Besides, as a result of the misalignment between confidence score and localization quality, it's hard to discriminate high-quality pseudo-labels from noisy predictions using only confidence-based criterion. To that end, we further introduce an uncertainty-based filter to help select reliable pseudo boxes for the above mixing operation. To the best of our knowledge, this is the first unified SSL framework for monocular 3D object detection. Mix-Teaching consistently improves MonoFlex and GUPNet by significant margins under various labeling ratios on KITTI dataset. For example, our method achieves around +6.34\% AP@0.7 improvement against the GUPNet baseline on validation set when using only 10\% labeled data. Besides, by leveraging full training set and the additional 48K raw images of KITTI, it can further improve the MonoFlex by +4.65\% improvement on AP@0.7 for car detection, reaching 18.54\% AP@0.7, which ranks the 1st place among all monocular based methods on KITTI test leaderboard. The code and pretrained models will be released at \href{https://github.com/yanglei18/Mix-Teaching}{here}. 
		
	\end{abstract}
	
	\section{Introduction}
	Monocular 3D object detection is the task of predicting the categories and 3D bounding boxes for surrounding objects with a single image. Owing to its distinct advantages and potential applications in autonomous driving and robotics, this task has attracted extensive attention of researchers from both academia and industry. In recent years, many innovative detectors have emerged and achieved increasing accuracy. However, most of these methods are heavily dependent on labeled data. Compared with the human-annotated images that are often expensive and time-consuming, raw images are  easier to achieve large-scale collection. Thus, taking full advantage of both labeled and unlabeled data in model training is a promising approach to alleviate the heavy reliance on human annotations.
	
	\begin{figure}[t]
		\centering
		\includegraphics[width=4.1cm]{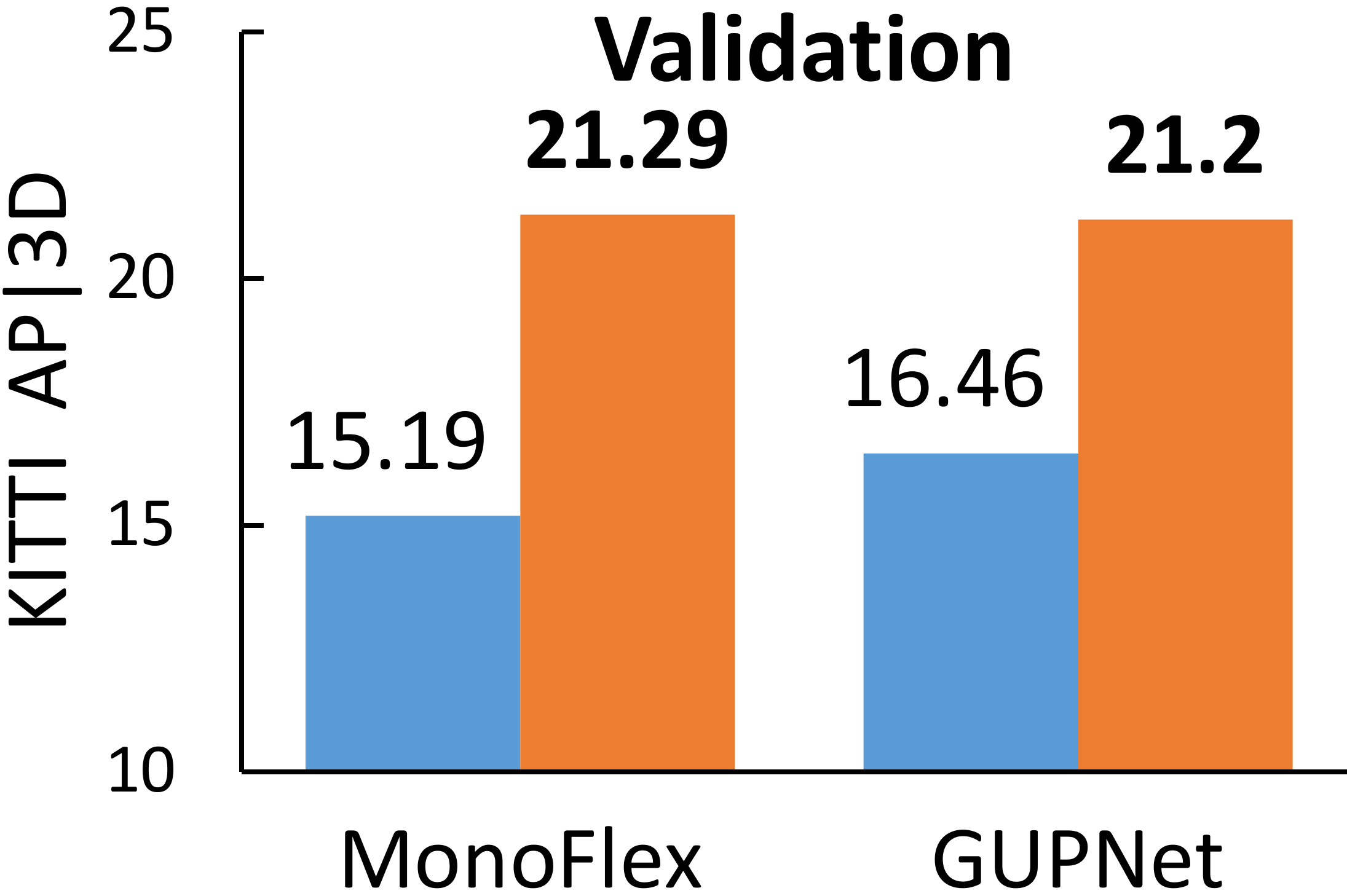}
		\includegraphics[width=4.1cm]{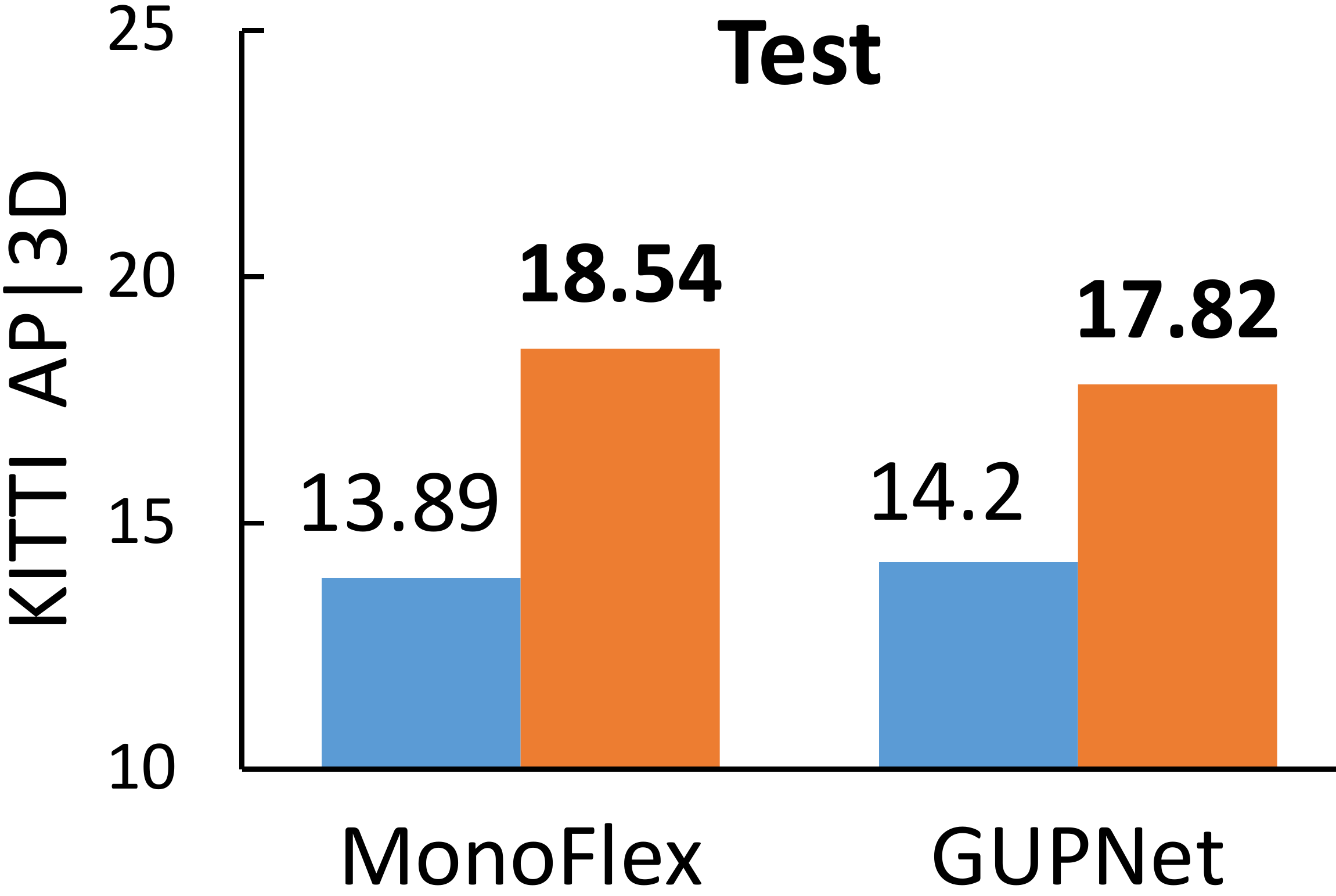}
		\caption{\textbf{Performance comparison.} The fully-supervised baselines are represented in {\color{cyan}{cyan}}, and their semi-supervised training results with our Mix-Teaching are displayed in {\color{orange}{orange}}. Our proposed method outperforms the baselines by a large margin on both KITTI validation and test set for Car category.}
		\label{fig:one}
	\end{figure}
	
	Semi-supervised learning (SSL) can help effectively improve the performance of fully-supervised baselines by employing both labeled and unlabeled data. In recent years, plentiful SSL methods for classification\cite{Zhang2021FlexMatchBS,Xie2020SelfTrainingWN,Berthelot2020ReMixMatchSL,Sohn2020FixMatchSS}, 2D object detection\cite{Zhou2021InstantTeachingAE,Tang2021HumbleTT,Jeong2019ConsistencybasedSL,Jeong2021InterpolationbasedSL,Wang2021DataUncertaintyGM,Yang2021InteractiveSW,Xu2021EndtoEndSO,Li2021RethinkingPL,sohn2020detection,Zhang2021SemiSupervisedOD} and LiDAR-based 3D object detection\cite{Wang20213DIoUMatchLI,Zhao2020SESSSS} have been proposed and applied.  It can be generally divided into pseudo labeling and consistency regularization. Pseudo labeling first generates pseudo labels for unlabeled data by self-training\cite{Xie2020SelfTrainingWN} or Mean Teacher\cite{Tarvainen2017MeanTA}. Then the student model is trained to predict the same pseudo labels on the same unlabeled images applied with label-preserving data augmentations. In this way, the student model can learn useful information from pseudo labels. Consistency regularization adds a consistency loss to enforce the model make stable predictions on different disturbed data, which helps improving model generalization ability. But as far as we know, there hardly exists a semi-supervised learning framework specially designed for monocular 3D object detection.
	
	Due to the challenge of recovering depth information from a single image, the accuracy of monocular 3D object detection is lagging significantly behind that of 2D object detection and LiDAR-based 3D object detection. Take KITTI \cite{geiger2012we} benchmark for example, the state of the art methods for monocular 3D object detection just achieve less than 15\% AP@0.7, while the candidates for the other two tasks have reached more than 85\%-96\% AP@0.7. This means that the pseudo labels for unlabeled images are predicted by image-based 3D object detectors with lower precision and recall. Lower precision signifies that more incorrect predictions are possible to be used as labels for unlabeled samples, which can lead to serious confirmation bias. On the other hand, lower recall explains the true positive labels on each image are far from enough to provide adequate supervision signals. Meanwhile, lacking of pseudo labels for plentiful objects can further cause miss-detection. However, most existing SSL methods directly employ the original unlabeled images or the mixup of two images as in Instant-Teaching\cite{Zhou2021InstantTeachingAE}, which can't handle the situations imposed by low recall pseudo labels effectively. To overcome these issues, we propose Mix-Teaching, an general semi-supervised learning framework for most monocular 3D object detectors.
	
	One key challenge for semi-supervised monocular 3D object detection is the extremely low recall of pseudo-labels. For an unlabeled image, only a small part of the objects on this image are correctly detected (less true positive), whereas most of the remaining instances are ignored (more false nagetive). Sparsely distributed true positives fail to provide adequate supervision signals in semi-supervised training. Meanwhile, training data with overmuch missing labels tends to bring miss-detection to most monocular 3D object detectors. In the proposed Mix-Teaching, we firstly predict pseudo-labels for unlabeled data by self-training. Unlabeled samples are then split into image patches collection with high-quality pseudo labels and the collection of background images containing no objects. Subsequently, the student model is trained on the mixed images that are created by merging the above instance image patches into empty backgrounds or human labeled images through strong data augmentation. In this way, the generated images are full of instances with high-quality pseudo labels while successfully avoiding the missing label cases, which is more effective for semi-supervised training. Finally, we adopt multi-stage training scheme to progressively propagate information from the labeled to the unlabeled data.
	
	Another key challenge for semi-supervised monocular 3D object detection is the confirmation bias. In other words, the model is overfitting to incorrect pseudo labels, which is caused by the extremely low precision of image-based 3D object detectors. Considering the misalignment between confidence score and localization quality, it's not exhaustive to eliminate incorrect labels using only confidence-based filter. To this end, we further propose an uncertainty-based filter to help remove noisy pseudo labels. In this method, the predictions by the models with identical structure but different parameters are used to estimate the uncertainty to each object. For the prediction set belonging to the same object, the higher uncertainty, the less predictions in this set, and the larger localization misalignment among them. We build a formula to represent the localization uncertainty in 3D object detection task. Base on both confidence-based filter and uncertainty-based filters, we manage to remove incorrect pseudo labels more effectively in semi-supervised training and thus alleviate the confirmation bias. Because the process of removing noisy pseudo labels is only carried out at the beginning of each training stage, the efficiency influence from uncertainty calculation is inappreciable.
	
	We benchmark Mix-teaching with SSL setting using the full KITTI\cite{geiger2012we} object data and KITTI\cite{geiger2012we} raw data. When using MonoFlex\cite{Zhang2021ObjectsAD} as backbone detector, Mix-Teaching achieve state of the art results on KITTI test leaderboard, which even surpasses the LPCG\cite{Peng2021LidarPC} method that directly relies on LiDAR-based 3D object detectors to generate pseudo labels. Furthermore, we provide the SSL experiments under different labling ratio, which can serve an initial baseline for semi-supervised monocular 3D object detection.
	
	Our contributions can be summarized as follows:
	\begin{itemize}
		\item We clarify the main difficulties in accomplishing semi-supervised learning for monocular 3D object detection and explain why existing SSL approaches can't handle these issues. On this basis, we propose Mix-Teaching, a general semi-supervised framework for monocular 3D object detection.
		\item To alleviate the confirmation bias, we further propose an uncertainty-based filter to help remove noisy pseudo labels effectively.
		\item Extensive experiments on KITTI dataset demonstrate the significant efficacy of Mix-Teaching framework. As the first study of SSL for monocular 3D object detection, this can serve as a crucial baseline for further researches.
	\end{itemize}
	
	\section{Related Work}
	
	\noindent {\bf Monocular 3D Object Detection.}
	A number of methods have been proposed for monocular 3D object detection. How to reconstruct spatial information more effectively is the core problem of these approaches. The Pseudo-LiDAR-based methods\cite{wang2019pseudo,you2019pseudo,Ma2019AccurateM3,Ma2020RethinkingPR,Ding2020LearningDC} firstly transform the input image to dense artificial point clouds with the existing depth estimation algorithms\cite{Daz2019SoftLF,qiao2020vip} and then employs LiDAR-based 3D object detectors\cite{Lang2019PointPillarsFE,Zheng2021SESSDSS}. The geometry-based methods\cite{mousavian20173d,Li2019GS3DAE,Liu2019DeepFD} infer depth information based on the 2D/3D geometry constraint of specific reference. Another keypoint-based works\cite{Zhou2019ObjectsAP,Liu2020SMOKESM,Zhang2021ObjectsAD,Lu2021GeometryUP,Chen2020MonoPairM3,Ma2021DelvingIL,Chen2021MonoRUnM3,Reading2021CategoricalDD} directly estimate the 3D properties of instance relying on the high-dimensional features at keypoint position.
	
	\noindent {\bf Semi-supervised Learning.}
	Semi-supervised Learning focuses on training models with both labeled and unlabeled data, which has achieved state-of-the-art performance on classification\cite{Zhang2021FlexMatchBS,Xie2020SelfTrainingWN,Berthelot2020ReMixMatchSL,Sohn2020FixMatchSS}, 2D object detection\cite{Zhou2021InstantTeachingAE,Tang2021HumbleTT,Jeong2019ConsistencybasedSL,Jeong2021InterpolationbasedSL,Wang2021DataUncertaintyGM,Yang2021InteractiveSW,Xu2021EndtoEndSO,Li2021RethinkingPL,sohn2020detection,Zhang2021SemiSupervisedOD} and LiDAR-based 3D object detection\cite{Wang20213DIoUMatchLI,Zhao2020SESSSS}. One popular type of SSL is consistency regularization, which constrains the outputs of different augmented inputs to be consistent. CSD\cite{Jeong2019ConsistencybasedSL} is a consistency-based method for 2D object detection. This approach ensures the consistent predictions between input images and their flipped versions on both labeled and unlabeled data. SESS\cite{Zhao2020SESSSS} is a semi-supervised learning framework for LiDAR-based 3D object detection.  To enhance the model generalization ability, this method applies three consistency losses on two sets of 3D proposals from teacher and student networks. The other kind of SSL is pseudo labeling, which is based on high-quality pseudo labels and can be seen as the hard version of consistency regularization. FixMatch\cite{Sohn2020FixMatchSS} first generates pseudo labels on weakly augmented unlabeled images, and then the student model is trained to predict the same classifications on strong augmented data. Unbiased Teacher\cite{Liu2021UnbiasedTF} addresses the pseudo-labeling bias issue caused by class imbalance in 2D annotations with the help of EMA\cite{Tarvainen2017MeanTA} training and focal loss\cite{lin2017focal}. 3DIoUMatch\cite{Wang20213DIoUMatchLI} achieves semi-supervised 3D object detection in point cloud with a teacher-student mutual learning framework. To improve the quality of pseudo labels, all the predictions that fail to pass the thresholds on classification score, objectness confidence and 3D IoU will be filtered out.
	
	In spite of the success of SSL in classification, 2D object detection and LiDAR-based 3D object detection, there hardly exists a general semi-supervised learning framework specialized for monocular 3D object detection.
	
	\begin{figure*}
		\centering
		\includegraphics[width=1.0\textwidth]{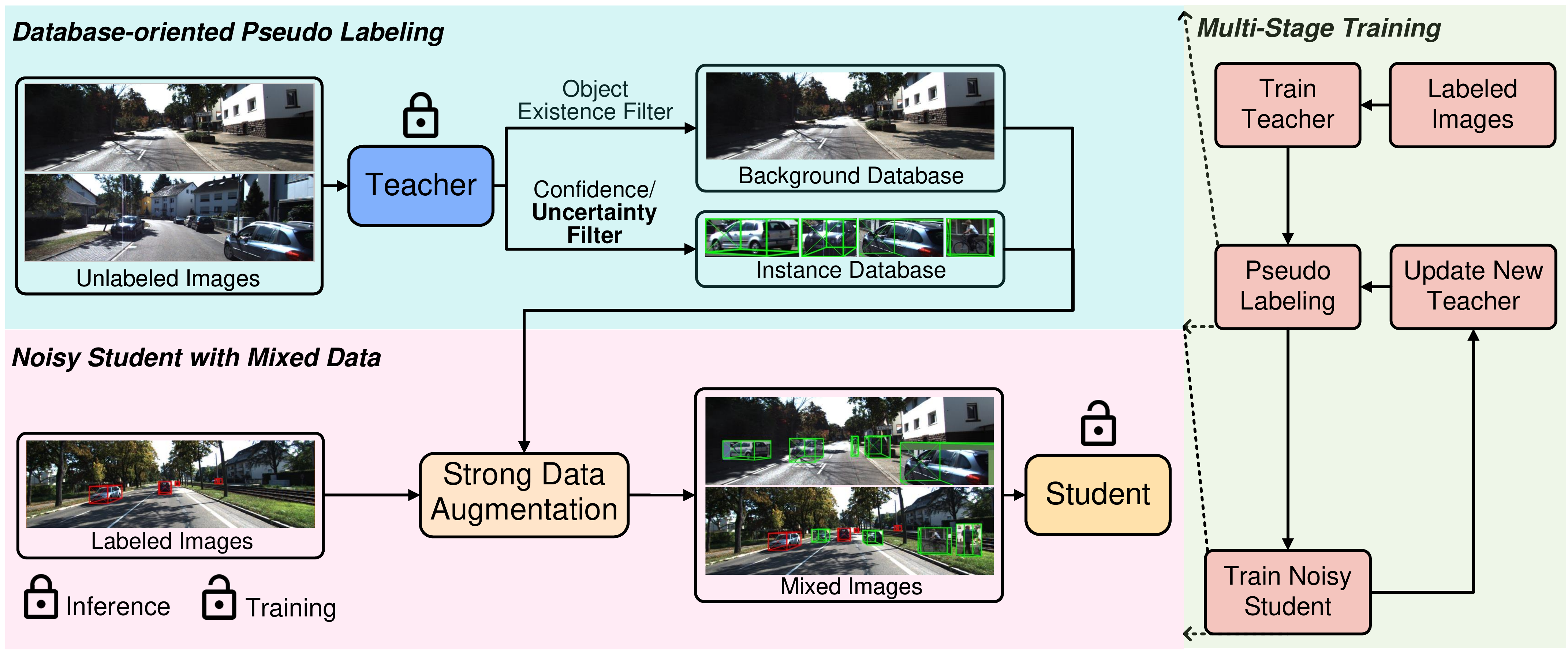}
		\caption{{\bf Overview of Mix-Teaching Framework.} Mix-Teaching follows multi-stage training scheme. There are two crucial process in each training stage. {\bf Database-oriented Pseudo Labeling:} Based on the pseudo labels that are generated by applying the teacher model on unlabeled images, We create two databases: one is background database that consists of images without any objects, the other one is instance database composed of image patches with high-quality pseudo labels. {\bf Noisy Student with Mixed Data:} The student model is trained on the mixed images that possess much more intensive and precise labels by merging image patches from the above instance database to labeled images or the ones from background database. See Section~\ref{sec:multi_stage_training} and Section~\ref{sec:mix-teaching_framework} for more details.}
		\label{fig:one}
	\end{figure*}
	
	\section{Method}
	In this section, we first give a mathematical definition of semi-supervised monocular 3D object detection task (see Section~\ref{sec:problem_definition}). Then, we show an overview of our training schema (see Section~\ref{sec:multi_stage_training}) and Mix-Teaching framework (see Section~\ref{sec:mix-teaching_framework}). The uncertainty-based filter is introduced in Section~\ref{sec:uncertainty_filter}.
	
	\subsection{Problem Definition}
	\label{sec:problem_definition}
	In semi-supervised monocular 3D object detection, we have labeled data 
	$I^L=\left\{(x_1^L,y_1^L ),…,(x_{n_l}^L,y_{n_l}^L )\right\}$ and abundant unlabeled data $I^U=\left\{x_1^U,…,x_{n_u}^U\right\}$, where $x$ is image, $y$ denotes the human-annotated label that contains category and 3D bounding box. $n_l$ and $n_u$ represent the number of labeled and unlabeled images respectively. We aim to significantly improve the performance of fully-supervised baselines by applying both labeled and unlabeled data in training.
	
	\subsection{Multi-stage Training Schema}
	\label{sec:multi_stage_training}
	We adopt a multi-stage training schema. The initial teacher model is trained on labeled data, followed by a pseudo-labeling process for the unlabeled data. Then we train a noisy student model using all the labeled and unlabeled images following decomposition and re-combination technique. This resulting model will be used as a new teacher model in the next stage.
	
	\subsection{Mix-Teaching Framework}
	\label{sec:mix-teaching_framework}
	We propose a SSL framework for monocular 3D object detection, called Mix-Teaching, as shown in Figure~\ref{fig:one}. This is a general approach that can be easily applied to most monocular 3D object detectors. Our Mix-Teaching is mainly composed of two stages: database-oriented pseudo-labeling and noisy student with mixed data.\\
	{\bf Database-oriented Pseudo Labeling.}
	To make the most of sparsely distributed high-quality pseudo labels in semi-supervised training, all the labels and background images need to be gathered together. As shown in Figure~\ref{fig:one}, we perform a test-time inference of the teacher model on unlabeled images to generate pseudo labels. By applying confidence-based and uncertainty-based filters, we create an instance database that is composed of instance-level image patches and their corresponding high-quality pseudo labels. Based on the object existence filter, we select all the background images that don't contain any predictions from unlabeled data and create a background database \\
	{\bf Noisy Student with Mixed Data.} 
	Based on the above two databases and labeled images, we create the mixed images containing more intensive and precise labels for semi-supervised training. There are two general strategies for this purpose. One way is to paste the image patches from the instance database on labeled images. Another way is to paste the instance-level patches on the images that come from the background database. During the process, the instance-level patches are pasted to target images according to their 2D bounding box on source images. To avoid over occlusion and other impossible outcomes, we additionally perform a 2D bounding box collision test to remove invalid paste operations.
	
\begin{figure*}
		\centering
		\includegraphics[width=1.0\textwidth]{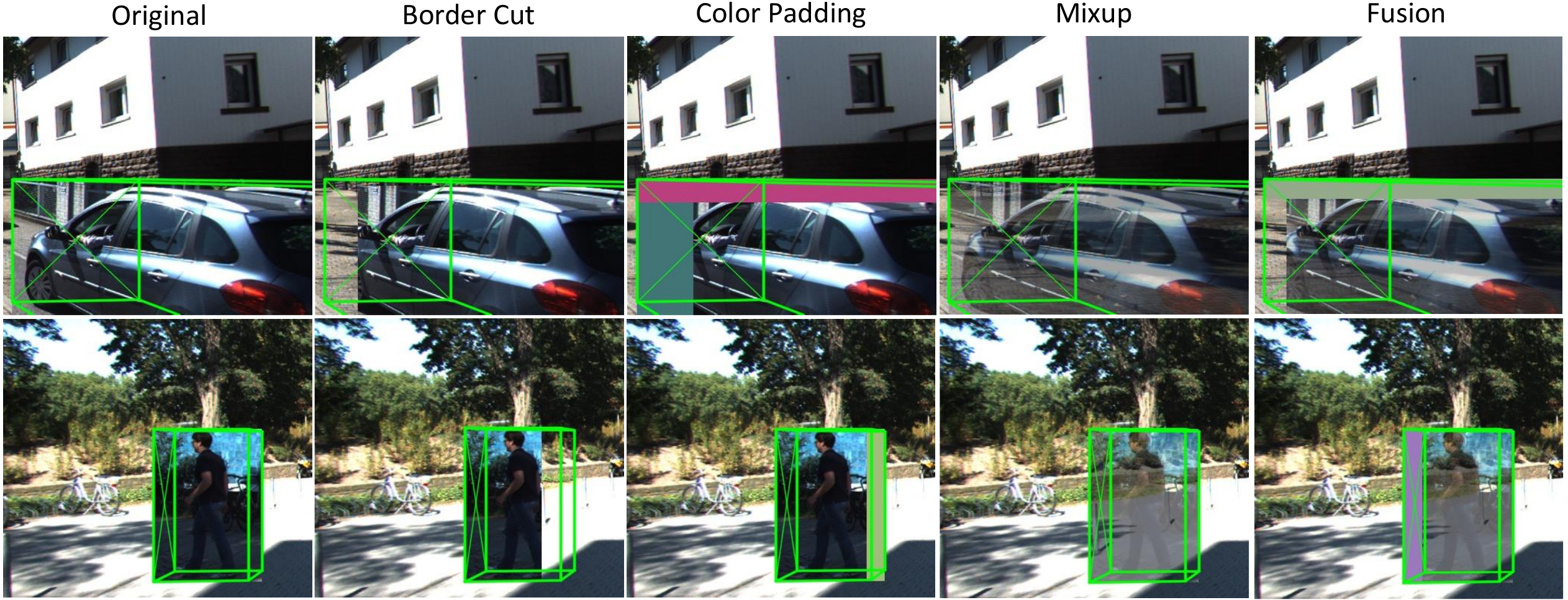}
		\caption{{\bf Visualization of box-level strong augmentations.}  From left to
			right: original image patch, border cut, color padding, mixup and the fusion of previous three methods.}
		\label{fig:two}
	\end{figure*}

	To alleviate the confirmation bias and improve the model generalization ability, we further propose a series of box-level strong data augmentations. For completeness, we describe the list of augmentations below. Each operation has a magnitude that decides the augmentation degree of strength. We visualize the augmented instances with single or fusion strategies mentioned above in Figure~\ref{fig:two}.
	\begin{enumerate}
		\item Border Cut (\textbf{B}): Before the pasting operation, cutting the horizontal or vertical border of image patch with a random ratio (0-0.3).
		\item Color Padding(\textbf{C}): Similar to the border cut, but replacing the cut operation with random color padding.
		\item Mixup(\textbf{M}): making a weighted average between the foreground image patches and backgrounds with a random ratio (0.6-1.0).
	\end{enumerate}
	
	When given a batch of mixed images and the corresponding human labels or pseudo labels, the model is trained by jointly minimizing the supervised loss and unsupervised loss as follows:
	\begin{equation}
		\mathcal{L} = \mathcal{L}_s + \lambda \times \mathcal{L}_u 
	\end{equation}
	where $\mathcal{L}$ is total loss, we use hyper-parameter $\lambda$ to balance the supervised loss $\mathcal{L}_s$ and the unsupervised loss $\mathcal{L}_u$.
	
	The supervised loss $\mathcal{L}_s$ consists of a classification loss $\mathcal{L}_{cls}$ and a regression loss $\mathcal{L}_{reg}$. It can be calculated as:
	\begin{equation}
		\mathcal{L}_s = \sum_{L}\frac{1}{N_l}\sum_{i}(\mathcal{L}_{cls}(b_l^i) + \mathcal{L}_{reg}(b_l^i))
	\end{equation}
	where $L$ denotes the index of labeled images in a batch, $N_l$ represents the number of human annotations for each image, $b_l^i$ is the $i$-th label in the $L$-th labeled image.
	
	The unsupervised loss $\mathcal{L}_u$ is computed on pseudo labels and can be written as:
	$$
	\mathcal{L}_u = \sum_{L}\frac{1}{N_u}\sum_{i}(\mathcal{L}_{cls}(b_u^i) + \mathcal{L}_{reg}(b_u^i))
	$$
	\begin{equation}
		+ \sum_{B}\frac{1}{N_u}\sum_{i}(\mathcal{L}_{cls}(b_u^i) + \mathcal{L}_{reg}(b_u^i))
	\end{equation}
	where $B$ indicates the index of background images, $N_u$ is the number of pseudo labels on each image, $b_u^i$ represents the $i$-th pseudo label on a labeled or a background image.
	
	For monocular 3D object detection, the extremely low recall and precision of related detectors make it a great challenge to apply the existing semi-supervised methods for 2D object detection that focus more on false positives but ignore false negatives to this field. In the Mix-Teaching proposed above, following decomposition and recombination methodology, we collect all positive instances and merge them into backgrounds to create newly mixed images for semi-supervised training. The positive instances indicate object-level image patches with high-quality pseudo labels. The backgrounds denote empty unlabeled images or labeled data. The newly mixed images will possess high recall and less false positives at the same time, which is effective to solve the extremely low recall and confirmation bias challenges in semi-supervised monocular 3D object detection.
	
	\subsection{Uncertainty-based Filter}
	\label{sec:uncertainty_filter}
	
\begin{figure}
		\centering
		\includegraphics[width=0.5\textwidth]{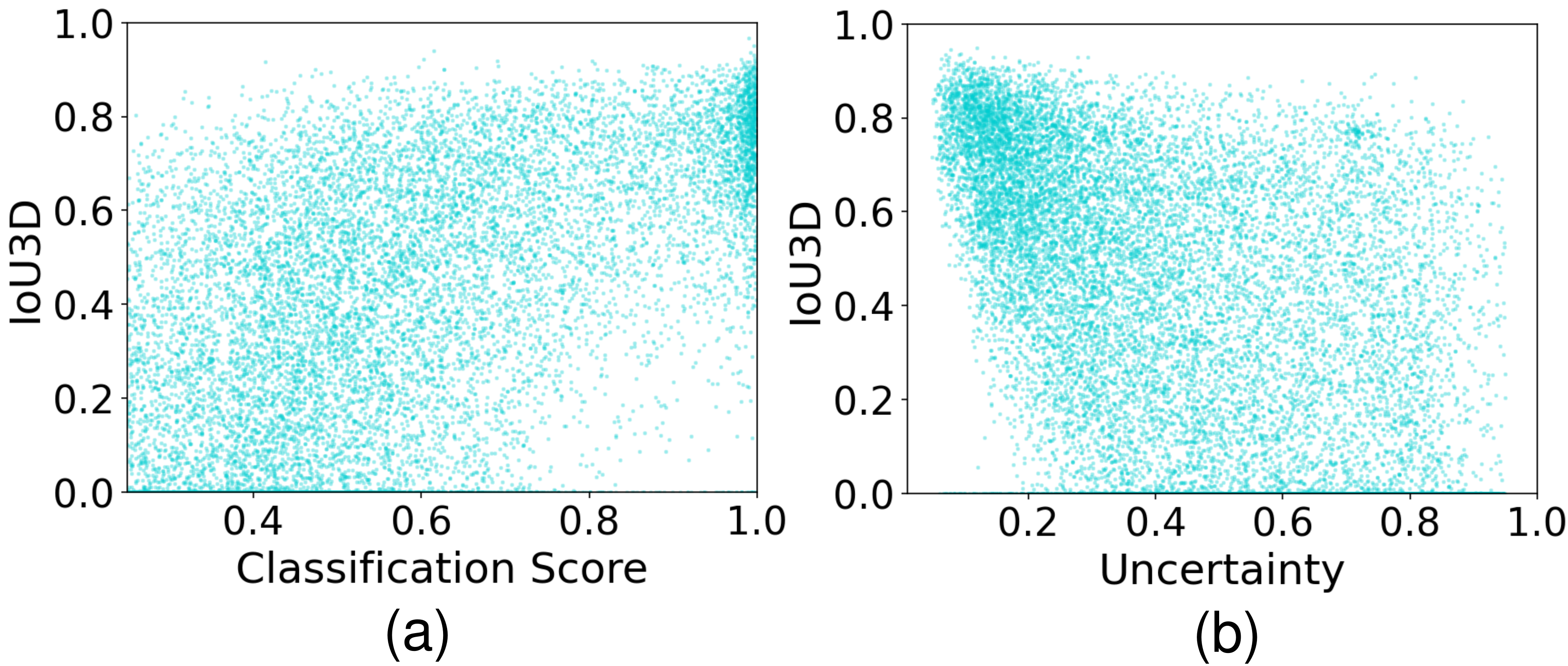}
		\caption{\textbf{The statistics of MonoFlex predictions on KITTI validation set.} (a) the relationship between the 3D IoU with ground truth box and classification score. (b) the relationship between the 3D IoU and localization uncertainty.}
		\label{fig:three}
	\end{figure}

	As shown in Figure~\ref{fig:three}(a), there exists a huge misalignment between the classification score and the localization precision of box candidates. A considerable proportion of predictions have a high confidence score but low 3D IoU with ground truth. When the high-quality pseudo labels are discriminate completely based on the confidence-based filter, a lot of incorrect pseudo labels will be used for semi-supervised training, which will strengthen the confirmation bias.
	
	In order to alleviate the above issues, it is necessary to remove noisy labels in semi-supervised training. To this end, we further propose an uncertainty-based filter in which we infer localization uncertainty on the basis of the discrepancy of the predictions for one object from multi models.
	
	When a certain image is given to $N$ isomorphic models with different parameters. For a specific object on this image, there will be $M$ predictions. We define the localization uncertainty mainly from two points of view: (1) the number of predictions $M$ associated with this object; (2) The discrepancy between these predicted boxes. The predictions number $M$ reflects the level of missed detection among $N$ models. The disparity in box candidates reveals the randomness in model predictions. 
	
	We calculate the uncertainty in the following steps:
	\begin{enumerate}
		\item All predictions from N models are stored in list $B$.
		\item Declare three lists $G$, $H$ and $U$. $G$ is used to store box clusters. Each cluster represents the predictions for a certain object from $N$ models .
		$H$ is for the box with highest confidence score in each cluster. $U$ saves the localization uncertainty for each box in list $H$. 
		\item Iterate through all the boxes in list $B$ to find the matching box that belongs to the current cluster $C$. The matching condition is defined as a box with a large overlap with the initial box $b_m$ of cluster $C$ under the condition $IoU3D > thr$. All matching boxes will be moved from list $B$ to cluster $C$. And then update the current cluster $C$ to list $G$.
		\item If there are still unprocessed boxes in list B, select the box $b_m$ that has the maximum score in list $B$ and move it to list $H$. Initialize a new cluster $C$ with box $b_m$ and proceed to step 3. 
		\item When all boxes in B are processed, calculate the uncertainty $u$ for each box cluster $C$ in list $G$ with the following equations. The results are added to list $U$.
		$$
		u = uncertain(C)=
		$$
		\begin{equation}
			1 - \frac{\sum_{i=0}^{M}\sum_{j=0}^{M}a_{ij}\times IoU3D(b_i, b_j)}{\sum_{i=0}^{N}\sum_{j=0}^{N}a_{ij}}
		\end{equation}
		
		\begin{equation}
			a_{ij} = \begin{cases} 
				1  & if \ i \neq j\\
				\beta & if \ i = j
			\end{cases}, 
		\end{equation}
		where $M$ is the number of boxes in cluster $C$, $N$ denotes the number of models, $b_i$ is the $i$-th box in cluster $C$, $a_{ij}$ represents the weight for each item. $\beta$ is the hyper-parameter that controls how the number of box candidates affects uncertainty.
	\end{enumerate}
	
	The uncertainty $u$ ranges from 0 to 1. When the value is 0, it indicates that there exists no miss-detection in $N$ models ($M=N$), and all $N$ box candidates are perfectly consistent. When the value is 1, it means that all models fail to detect this object.
	
	As shown in Figure~\ref{fig:three} (b), we visualize the relationship between the 3D IoU and localization uncertainty. Compared with the classification score, the obtained uncertainty can better measure the localization accuracy.
	
	Compared with 3D confidence\cite{Simonelli2021AreWM} or 3D IoU \cite{Wang20213DIoUMatchLI, Li2021Monocular3D} that requires related branch design for specific detectors, our uncertainty-based filter is model-independent and can be applied to many types of image-based 3D object detectors, which is more appropriate for the proposed general semi-supervised learning framework.
	
	\begin{table*}
		\centering
		\footnotesize
		\setlength\tabcolsep{3.5pt}
		\begin{tabular}{l|ccc|ccc|ccc}
			\toprule  
			\multirow{2}{*}{Method} & \multicolumn{3}{c|}{ 10\%}  &  \multicolumn{3}{c}{ 50\%} &  \multicolumn{3}{c}{ 100\%} \\ 
			~ &  Easy & Mod. & Hard & Easy & Mod. & Hard & Easy & Mod. & Hard\\ 
			\midrule
			GUPNet&   8.42 / 13.61& 5.14 / 8.67&  4.11 / 7.35&  18.21 / 26.28& 14.14 / 19.38& 11.12 / 16.63&  22.76 / 31.07& 16.46 / 22.94& 13.72 / 19.75 \\ 
			Ours &    16.55 / 22.57&  11.48 / 15.90&  9.44 / 13.13&  25.67 / 34.87& 19.76 / 26.19&  17.07 / 21.76&  29.12 / 38.48& 21.04 / 28.25&  17.56 / 24.36 \\ 
			Abs. Imp. &  +8.13 / 8.96&  +6.34 / +7.23&  +5.33 / +5.78&  +7.46 / +8.59& +5.62 / +6.91&  +4.95 / +5.13&  +6.36 / +7.41&  +4.58 / +5.31&  +3.84 / +4.61\\ 
			\midrule
			MonoFlex&  5.76 / 9.93&  4.67 / 7.68&  3.54 / 6.09&  21.91 / 28.24& 15.43 / 20.51&  13.09 / 18.37&  23.64 / 29.86& 17.51 / 23.05&  14.83 / 20.68 \\ 
			Ours &  14.43 / 18.80&  10.65 / 14.28&  8.41 / 11.73&  29.34 / 36.23& 20.63 / 26.70&  17.31 / 23.07&  29.74 / 37.45&  22.27 / 28.99& 19.04 / 25.31 \\ 
			Abs. Imp.&  +8.57 / +8.87&  +5.98 / +6.60&  +4.87 / +5.64&  +7.43 / +7.99& +5.20 / +6.19&  +4.22 / +4.70&  +6.10 / +7.56& +4.76 / +5.94& +4.21 / +4.63 \\ 
			\bottomrule 
		\end{tabular}
		\centering
		\caption{{\bf Quantitative results of AP$_{3D}$/AP$_{BEV}$(IoU=0.7)$|\scriptstyle R_{40}$ on KITTI val set under under different ratios of training set.} ``Abs. Imp." represents absolute improvements.}
		\label{tab:one}
	\end{table*}

	\begin{table*}
		\centering
		\small
		\setlength\tabcolsep{7.8pt}
		\begin{tabular}{l|c|c|c|ccc|ccc}
			\toprule[1pt]
			\multirow{2}{*}{Method} &  \multirow{2}{*}{Reference}& \multirow{2}{*}{GPU} &
			{Runtime} & \multicolumn{3}{c|}{AP$_{3D}$(IoU=0.7)$|\scriptstyle R_{40}$} & \multicolumn{3}{c}{AP$_{BEV}$(IoU=0.7)$|\scriptstyle R_{40}$}  \\ 
			& & & (ms)& Easy  & Mod. & Hard & Easy & Mod. & Hard \\ 
			\midrule[0.5pt]
			MonoGRNet\cite{Qin2021MonoGRNetAG} & TPAMI 2021& Tesla P40& 60 &  $9.61$ & $5.74$ & $4.25$ & $18.19$ & $11.17$ & $8.73$ \\
			MonoPair\cite{Chen2020MonoPairM3} & CVPR 2020& Tesla V100& 60 & $13.04$ & $9.99$ & $8.65$ 	& $19.28$ & $14.83$ & $12.89$ \\ 
			RTM3D\cite{Li2020RTM3DRM} & ECCV 2020& 1080Ti& 50 & $14.41$ & $10.34$ & $8.77$ & $19.17$ & $14.20$ & $11.99$ \\ 
			D$^4$LCN\cite{Ding2020LearningDC} & CVPR 2020& 1080Ti & 200 & $16.65$ & $11.72$ & $9.51$ & $22.51$ & $16.02$ & $12.55$ \\ 
			Monodle\cite{Ma2021DelvingIL} & CVPR 2021& 1080Ti & 40 & 17.23 & 12.26 & 10.29 & 24.79 & 18.89 & 16.00 \\
			MonoRUn\cite{Chen2021MonoRUnM3} & CVPR 2021 & 1080Ti& 70 & $19.65$ & $12.30$ & $10.58$ &  
			$27.94$ & $17.34$ & $15.24$ \\
			GrooMeD-NMS \cite{Kumar2021GrooMeDNMSGM} & CVPR 2021& Titian X & 120 & $18.10$ & $12.32$ & $9.65$ & $26.19$ & $18.27$ & $14.05$\\
			MonoRCNN\cite{Shi2021GeometrybasedDD} & ICCV 2021& -& 70 & $18.36$ & $12.65$ & $10.03$  & $25.48$ & $18.11$ & $14.10$   \\ 
			DDMP-3D\cite{Wang2021DepthconditionedDM}\textbf{$\ast$}& CVPR 2021& -& 180& 19.71& 12.78& 9.80& 28.08& 17.89& 13.44 \\
			Ground-Aware\cite{Liu2021GroundAwareM3} & RAL 2021& 1080Ti & 50 & $21.65$ & $13.25$ & $9/91$ & $29.81$ & $17.98$ & $13.08$ \\
			PCT\cite{Wang2021ProgressiveCT} & NIPS 2021& -& 487& 21.00 & 13.37 & 11.31 & 29.65 & 19.03 & 15.92 \\
			CaDDN\cite{reading2021categorical} & CVPR2021& 2080Ti & 485 &	19.17 &	13.41 &	11.46 &	27.94 &	18.91 &	17.19\\
			DFR-Net\cite{Zou2021TheDI}\textbf{$\ast$} & ICCV 2021& -& 180 & $19.40$ & $13.63$ & $10.35$  & $28.17$ & $19.17$ & $14.84$   \\ 
			
			MonoEF\cite{Zhou2021Monocular3O} & TPAMI 2021& -& 30 & $21.29$ & $13.87$ & $11.71$ & $29.03$ & $19.70$ & $17.26$ \\
			MonoFlex\cite{Zhang2021ObjectsAD}$^\dagger$ & CVPR 2021& 2080Ti & 30 & $19.94$ & $13.89$ & $12.07$  & $28.23$ & $19.75$ & $16.89$ \\ 
			AutoShape\cite{Liu2021AutoShapeRS} & ICCV 2021&	2080Ti & 52 &	22.47 &	14.17 &	11.36 &	30.66 &	20.08 &	15.95\\
			GUPNet\cite{Lu2021GeometryUP}$^\dagger$ & ICCV 2021& 2080Ti & 26 & $20.11$ & $14.20$ & $11.77$ & -& -& -  \\
			MonoDTR\cite{Huang2022MonoDTRM3} & CVPR 2022& -& 37& 21.99& 15.39& 12.73& 28.59& 20.38& 17.14 \\
			MonoDETR\cite{Zhang2022MonoDETRDT} & CVPR 2022& -& 40& \textbf{23.65}& 15.92& 12.99& \underline{32.08}& 21.44& 17.85 \\
			MonoDistill\cite{Chong2022MonoDistillLS}\textbf{$\ast$} & ICLR 2022& 1080 Ti& 40& 22.97& 16.03& 13.60& 31.87& 22.59&  19.72 \\
			MonoJSG\cite{Lian2022MonoJSGJS} & CVPR 2022& -& 42& 24.69&  16.14& 13.64& 32.59& 21.26& 18.18 \\
			DD3D\cite{Park2021IsPN}\textbf{$\ast$} & ICCV 2021& 2080Ti & 60 & 23.19 & 16.87 & 14.36 & 32.35 & 23.41 & 20.42\\
			LPCG\cite{Peng2021LidarPC}\textbf{$\ast$}  ~& Arxiv 2021& 2080Ti & 30 & 25.56 & 17.80 & \underline{15.38} & \underline{35.96} & \textbf{24.81} & \textbf{21.86} \\
			\midrule[0.5pt]
			\textbf{GUPNet + Ours} & -& 2080Ti & 26 & \textbf{{27.55}} & \underline{17.82} & {14.72}  & \textbf{{36.39}} & 24.14 & \underline{20.49} \\ 
			Abs. Imp. & -& -& -&  $+7.44$ & $+3.62$ & $+2.95$  & - & - & - \\ 
			\midrule[0.5pt]
			\textbf{MonoFlex + Ours} & -& 2080Ti & 30 & \underline{26.89} & \textbf{18.54} & \textbf{{15.79}}  & 35.74 & \underline{24.23} & \underline{20.80} \\ 
			Abs. Imp. & -& -& -& $+6.95$ & $+4.65$ & $+3.72$  & $+7.51$ & $+4.48$ & $+3.91$ \\ 
			\bottomrule[1pt]
		\end{tabular}
		\centering
		\caption{\textbf{Performance of the Car category on KITTI test set}. We use {\textbf{bold}} to highlight the highest results and {\underline{underlined}} for the second-highest ones. $^\dagger$ represents the baseline we employed. All methods are ranked by $AP_{3D}$ on moderate setting (same as KITTI leaderboard), Our method outperforms the baseline by a large margin and achieves the best performance.}
		\label{tab:two}
	\end{table*}
	
	\section{Experiments}
	\subsection{Dataset and Metrics}
	Contrasted with nuscenes\cite{Caesar2020nuScenesAM} and waymo\cite{Sun2020ScalabilityIP} dataset that lacks of a large amount of unlabeled data, KITTI\cite{geiger2012we} dataset provides 15K frames labeled data and 48K unlabeled images, which is more appropriate for semi-supervised learning research that relies on limited labeled data and larger scale unlabeled images. Therefore, we evaluate our Mix-teaching on the challenging KITTI\cite{geiger2012we} dataset. KITTI contains 7,481 images for training and 7,518 images for testing. Since we have no access to the manual annotations of testing set, the training set is further split into 3,712 training samples and 3,769 validation samples as mentioned in \cite{Chen20153DOP} for local evaluation. Besides, there are additional raw data consists of 48K temporal images. These images don't coincide with the above training or testing set, and thus can be used as unlabeled data for semi-supervised training. We use the average precision on Car, Pedestrian and Cyclist for 3D and bird’s eye view (BEV) object detection as the metrics. Following \cite{Simonelli2019DisentanglingM3}, all the evaluation results on validation and testing set are based on $AP_{40}$ instead of the original 11-point interpolated average precision.
	
	\subsection{Implement Details}
	We adopt the MonoFlex\cite{Zhang2021ObjectsAD} and GUPNet\cite{Lu2021GeometryUP} as two baseline detectors. The localization uncertainty is calculated based on five models from different training rounds. During the pseudo labeling process, only the predictions with confidence score larger than 0.7 and localization uncertainty less than 0.25 will be added to the instance database. The images without any detections are collected to build background database. During the student model training period, 
	We initialize the student with the previous teacher model. the images from background database are selected with a chance of 42\% apart from labeled images. For the box-level strong data augmentation, we apply mixup on every instance image patches. border cut and color padding augmentation are employed with a chance of 50\%.  We set the hyper-parameter $ \lambda=1.0$. Following the multi-training scheme, we conduct three cycles of semi-supervised training for all experiments. 
	
	\subsection{Quantitative Results}
	\noindent {\bf Comparison with Fully-supervised Baselines.} We make a detailed comparison with the supervised baselines, including GUPNet\cite{Lu2021GeometryUP} and MonoFlex\cite{Zhang2021ObjectsAD}, under different ratios of training set. All 48K raw images of KITTI are used as unlabeled data for semi-supervised training. As depicted in Table ~\ref{tab:one}, Mix-Teaching significantly outperforms MonoFlex\cite{Zhang2021ObjectsAD} and GUPNet\cite{Lu2021GeometryUP} under each ratio settings, which verify the effectiveness of our semi-supervised framework. when using only 10\% labeled data, our approach gains around +6.34\% and +5.98\% $AP_{3D}$ improvements on moderate level over MonoFlex and GUPNet baselines. This indicates our framework is able to learn knowledge from unlabeled data, and the effect is more obvious when the number of labeled data is scarce. Furthermore, it is worth pointing out that when using all training set, our Mix-Teaching is able to further outperforms the upper-bound performance of two baselines by a large margin.
	
	\noindent {\bf Results on KITTI Test Set.} 
	We evaluate the proposed Mix-Teaching on KITT test set using MonoFlex\cite{Zhang2021ObjectsAD} and GUPNet\cite{Lu2021GeometryUP} as two base monocular detectors. Table~\ref{tab:two} shows the quantitative results of our method and other top performance detectors from the official KITTI leaderboard. Overall, Mix-Teaching achieves superior results over all two baselines across all settings under fair conditions. For instance, the proposed method improve the $AP_{3D}$ of GUPNet\cite{Lu2021GeometryUP} by \textbf{+7.44/+3.62/+2.95} absolute improvements under easy/moderate/hard setting. Meanwhile, our approach increases the same metric of MonoFLex\cite{Zhang2021ObjectsAD} from 19.94/13.89/12.07 to 26.89/18.54/15.79, which is absolutely remarkable. What's more, Our Mix-Teaching even surpasses the LPCG\cite{Peng2021LidarPC} that directly relies on LiDAR-based 3D object detectors to help generate pseudo labels on $AP_{3D}$ metric using the same MonoFlex baseline. We rank the 1st place according to $AP_{3D}$ on moderate setting (same as KITTI leaderboard).
	
	\noindent {\bf Results on Pedestrian and Cyclist Categories.}
	Compared with Car, Pedestrian and Cyclist are more challenging to be detected owing to their non-rigid structure, small scale. As shown in Table {\color{red}3}, our Mix-teaching can further boost the AP$_{3D}$(IoU=0.5)$|\scriptstyle R_{40}$ metric of MonoFlex\cite{Zhang2021ObjectsAD} baseline around 18\% relative improvements for pedestrian and 108\% for cyclist on the test set, which demonstrates the effectiveness of our method on small-scale objects.
	
	\begin{table}[h!t]
		\setlength\tabcolsep{1.5pt}
		\centering
		\footnotesize
		\begin{tabular}{l|c|ccc|ccc}
			\toprule   
			\multirow{2}{*}{Method} & \multirow{2}{*}{Cat.}  & \multicolumn{3}{c|}{AP$_{3D}|\scriptstyle R_{40}$}& \multicolumn{3}{c}{AP$_{BEV}|\scriptstyle R_{40}$} \\
			~ & ~& Easy & Mod. & Hard & Easy & Mod. & Hard  \\ 
			\midrule         
			MonoFlex & \multirow{3}{*}{Ped.}& 9.43 & 6.31 & 5.26 & 10.36 & 7.36 & 6.29\\
			Ours  &~ &  \bf 11.67 & \bf 7.47 & \bf 6.61 & \bf 12.34 & \bf 8.40 & \bf 7.06\\
			Rel. Imp.(\%)& ~ &  23.75 $\uparrow$ & 18.38 $\uparrow$ & 25.67 $\uparrow$ & 19.11 $\uparrow$ & 14.13 $\uparrow$ & 12.24 $\uparrow$ \\
			\midrule         
			MonoFlex &  \multirow{3}{*}{Cycl.} & 4.17 & 2.35 & 2.04 & 4.41 & 2.67 & 2.50\\
			Ours  &~ &  \bf 8.04 & \bf 4.91 & \bf 4.15 & \bf 8.56 & \bf 5.36 & \bf 4.62\\
			Rel. Imp.(\%)& ~ & 92.81 $\uparrow$ & 108.94 $\uparrow$ & 103.43 $\uparrow$ & 94.10 $\uparrow$ & 100.75 $\uparrow$ & 84.80 $\uparrow$ \\
			\bottomrule 
		\end{tabular}
		\label{tab:three1}
		\centering
		\caption{\textbf{Quantitative results for Pedestrian and Cyclist on KITTI test set.}  ``Rel. Imp." represents relative improvements.}
	\end{table}
	
	\subsection{Ablation Studies}
	In this section, we perform ablation studies to investigate the effects of each elements. We use MonoFlex\cite{Zhang2021ObjectsAD} as the base detector. The training of ablation experiments is conducted on the full KITTI training set. The results for car category are evaluated on the corresponding validation set.
	
	\noindent {\bf The Scale of Unlabeled Data} We investigate if the proposed semi-supervised learning strategy keeps improving performance with increasing unlabeled data. As show in table~\ref{tab:four}, Compared with the results of 24K KITTI raw data, the experiment when using the whole 48K unlabeled data can further improve the AP$3D$ for car category from 20.61\% to 22.27\%. This means that, with more unlabeled images, our Mix-Teaching can improve the accuracy of fully supervised detectors to a new level.
	
	\noindent{\bf Background Database} Next, we investigate if the backbone database is necessary during the student model training period. As shown in table~\ref{tab:four}, when using half KITTI raw data, we gain +2.23\% and +2.45\% absolute improvement over the without background database version on AP$_{3D}$ and AP$_{BEV}$ respectively. And when it comes to all the unlabeled data condition, background database brings about significant improvements as well.
	
	\begin{table}[h!t]
		\centering
		\footnotesize
		\setlength\tabcolsep{3pt}
		\renewcommand\arraystretch{1.0}
		\begin{tabular}{c|c|ccc|ccc}
			\toprule  
			\multirow{2}{*}{Raw Data} & \multirow{2}{*}{Background} & \multicolumn{3}{c|}{AP$_{3D}|\scriptstyle R_{40}$} & \multicolumn{3}{c}{AP$_{BEV}|\scriptstyle R_{40}$} \\ 
			~ & ~ &  Easy & Mod. & Hard & Easy & Mod. & Hard\\ 
			\midrule
			-&  -& 23.64 & 17.51 & 14.83 & 29.86 & 23.05 & 20.68 \\
			\midrule
			\multirow{2}{*}{50\%} & -&  23.79& 18.29 & 15.66 & 32.55 & 24.15  & 21.58 \\
			~ & $\surd$ & 27.49 & 20.61 & 17.68 & 36.19 & 26.60 & 23.13 \\
			\midrule
			\multirow{2}{*}{100\%} & -& 24.22 & 19.02 & 16.16 & 33.33 & 25.70 & 22.34 \\
			~ & $\surd$ & \textbf{29.74} & \textbf{22.27} & \textbf{19.04} & \textbf{37.45} & \textbf{28.99} & \textbf{25.31} \\
			\bottomrule 
		\end{tabular}
		\caption{\textbf{Ablation study on the effects of background database and unlabeled data scale.}}
		\label{tab:four}
	\end{table}
	
	\noindent {\bf Box-level Data Augmentations.}
	We ablate the effects of box-level data augmentations. As shown in Table~\ref{tab:five}, all border cut, color padding and mixup are helpful to improve performance. The combination of the above three data augmentations can further improve the performance of Mix-Teaching.

	\begin{table}[h!t]
		\centering
		\footnotesize
		\setlength\tabcolsep{6pt}
		\renewcommand\arraystretch{1.0}
		\begin{tabular}{c|c|c|ccc|ccc}
			\toprule
			\multirow{2}{*}{B} & \multirow{2}{*}{C} & \multirow{2}{*}{M} &  \multicolumn{3}{c|}{AP$_{3D}|\scriptstyle R_{40}$} &  \multicolumn{3}{c}{AP$_{BEV}|\scriptstyle R_{40}$} \\ 
			~ & ~ & ~ & Easy & Mod. & Hard & Easy & Mod. & Hard \\
			\midrule
			- & - & - & 26.44 & 20.04 & 17.18 & 35.04 & 25.87 & 22.54 \\ 
			\midrule
			$\surd$  & - & - & 28.40 & 21.05 & 18.15 & 36.73 & 26.95 & 23.48 \\
			- & $\surd$  & - & 27.22 & 20.76 & 17.85 & 35.93 & 26.71 & 23.26 \\ 
			- & - & $\surd$ & 27.74 & 21.21 & 18.32 & 36.99 & 27.08 & 23.54 \\ 
			$\surd$ & $\surd$ & $\surd$ & \textbf{29.74} & \textbf{22.27} & \textbf{19.04} &  \textbf{37.45} & \textbf{28.99} & \textbf{25.31} \\
			\bottomrule 
		\end{tabular}
		\caption{\textbf{Ablation study on the effects of box-level data augmentations.} ``B" implies the border cut, ``C" denotes the color padding, ``M" represents the mixup.}
		\label{tab:five}
	\end{table}

	\noindent {\bf Confidence-based and Uncertainty-based Filters.} 
	As shown in Table~\ref{tab:six}, the results of applying uncertainty-based filter surpass that of only using confidence-based filter by 1.06\% AP$_{3D}$ on moderate set, which explains that our proposed uncertainty-filter is much more effective. When employing both of the two filters, we achieve the best results.
	
	\begin{table}[h!t]
		\centering
		\footnotesize
		\setlength\tabcolsep{6pt}
		\begin{tabular}{c|c|ccc|ccc}
			\toprule  
			\multirow{2}{*}{Conf.} & \multirow{2}{*}{Unc.} & \multicolumn{3}{c|}{AP$_{3D}|\scriptstyle R_{40}$} & \multicolumn{3}{c}{AP$_{BEV}|\scriptstyle R_{40}$} \\ 
			~ & ~ &  Easy & Mod. & Hard & Easy & Mod. & Hard\\ 
			\midrule
			- & - &  23.83&  17.81&  15.13&  30.52& 23.14 & 19.96 \\ 
			\midrule
			$\surd$ & - & 26.51 & 20.08 & 17.21 & 35.14 & 25.95 & 22.59 \\ 
			- &  $\surd$ & 27.78  & 21.14  & 18.18 & 35.36  & 26.87  & 23.52 \\
			$\surd$ &  $\surd$ & \textbf{29.74} & \textbf{22.27} & \textbf{19.04} & \textbf{37.45} & \textbf{28.99} & \textbf{25.31} \\
			\bottomrule 
		\end{tabular}
		\caption{\textbf{Ablation study on the effects of two filters.} ``Conf." denotes the filter based on the confidence. ``Unc." represents the uncertainty-based filter.}
		\label{tab:six}
	\end{table}
	
	\noindent {\bf The Thresholds for Confidence-based and Uncertainty-based Filters.} 
	We studies the effects of different confidence score thresholds and uncertainty thresholds in discriminating high-quality pseudo labels. As shown in Table~\ref{tab:seven}, The best performance is achieved when the confidence threshold is set to 0.7 and the uncertainty threshold is set to 0.25. 
	
	\begin{table}[h!t]
		\centering
		\footnotesize
		\setlength\tabcolsep{3pt}
		\begin{tabular}{c|c|ccc|ccc}
			\toprule  
			\multirow{2}{*}{Conf. Thre.} & \multirow{2}{*}{Unc. Thre.} & \multicolumn{3}{c|}{AP$_{3D}|\scriptstyle R_{40}$} & \multicolumn{3}{c}{AP$_{BEV}|\scriptstyle R_{40}$} \\ 
			~ & ~ &  Easy & Mod. & Hard & Easy & Mod.& Hard\\ 
			\midrule
			0.6 & 0.25 & 27.84 & 21.63 & 18.26 & 36.04& 26.51 & 23.49 \\ 
			0.7 & 0.25 & \textbf{29.74}  & \textbf{22.27}  & \textbf{19.04}& \textbf{37.45}& \textbf{28.99}  & \textbf{25.31} \\
			0.8 & 0.25 & 28.07 & 21.96 & 18.54 & 36.69& 27.06 & 23.90 \\
			0.9 & 0.25 & 26.31 & 20.20 & 16.59 & 33.59& 24.58 & 21.24 \\
			\midrule
			0.7 & 0.45 & 26.97  & 21.65 & 18.31 & 34.09& 26.50  & 23.45 \\
			0.7 & 0.35 & 28.89 & 21.98 & 18.51 & 36.75& 26.85 & 23.72 \\
			0.7 & 0.15 & 27.34 & 21.42 & 18.37 & 35.45& 26.32 & 23.50 \\
			\bottomrule 
		\end{tabular}
		\caption{\textbf{Ablation study on the effects of different uncertainty and confidence thresholds.} ``Conf. Threshold" denotes the threshold of confidence-based filter. ``Unc. Threshold" represents the threshold of uncertainty-based filter.}
		\label{tab:seven}
	\end{table}

	\begin{figure*}
		\centering
		\includegraphics[width=1.0\textwidth]{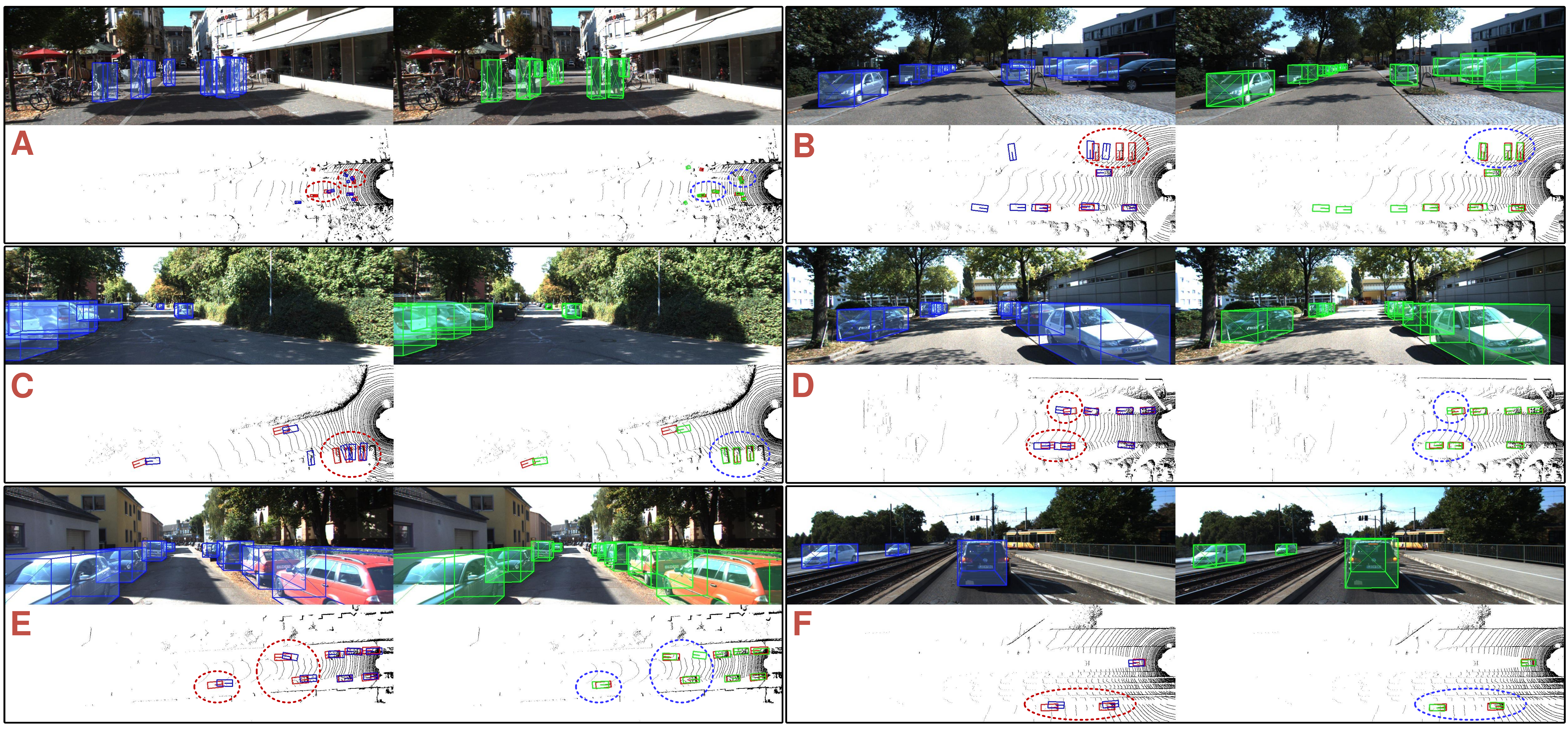}
		\caption{{\bf Qualitative results on the KITTI val set.} We present four pairs of comparisons marked with capital letters from A to F. Each pair consists of four pictures, the upper left displays the predictions of MonoFlex\cite{Zhang2021ObjectsAD} baseline({\color{blue}blue}), the lower left is its representation in the bird’s-eye view. The upper right shows the results of our Mix-Teaching({\color{green}green}), the lower right is its bird's-eye view display. The {\color{red}red} boxes in the bird’s eye view represent ground truths. We use dashed ovals to 
			highlight the pronounced difference in the predictions.}
		\label{fig:four}
	\end{figure*}
	
	\subsection{Qualitative Analysis}
	From the qualitative results shown in Figure~\ref{fig:four} , Mix-teaching can improve the performance of MonoFlex\cite{Zhang2021ObjectsAD} in various street scenes. As highlighted by the red ovals, our method can produce superior performance for person and cyclist(A in Figure~\ref{fig:four}), over-occluded objects (C in Figure~\ref{fig:four})  and cars in both close range (B-C in Figure~\ref{fig:four}) and long range (D,E,F in Figure~\ref{fig:four}), which demonstrates the efficiency of our Mix-teaching.

	\section{Conclusion and Future Work}
	In this paper, we proposed Mix-Teaching, a general semi-supervised learning framework for monocular 3D object detection. Our method first generates pseudo labels for unlabeled data by self-training. Then, following decomposition and re-combination technique, we break the limitation of original images and create newly diverse and label-rich mixed images for semi-supervised training, which can effectively handle the issues imposed by the extremely lower precision and recall of initial pseudo labels. With the proposed uncertainty-based filter, we manage to filter poorly positioned pseudo labels effectively, leading to less noise so as to alleviate confirmation bias. Experiments on KITTI dataset show that Mix-Teaching manages to improve the baseline model by a large margin under various labeling ratios. More importantly, when using 100\% training set and MonoFlex as backbone detector, we successfully rank the first place among all monocular 3D object detectors on KITTI test leaderboard. In this way, we can continuously boost monocular 3D object detectors by collecting more unlabeled images, which has great economic significance in autonomous driving. Moreover, the proposed Mix-Teaching follows the multi-stage training scheme. Adopting end-to-end training fashion will be left for future work.
	
	{\small
		\bibliographystyle{ieee_fullname}
		\bibliography{egbib}
	}
	
\end{document}